\documentclass[conference]{IEEEtran}
\IEEEoverridecommandlockouts

\usepackage{cite}
\usepackage{amsmath,amssymb,amsfonts}
\usepackage{algorithmic}
\usepackage{graphicx}
\usepackage{textcomp}
\usepackage{xcolor}
\usepackage{hyperref}
\hypersetup{hidelinks}
\usepackage{siunitx}
\usepackage[caption=false]{subfig}
\usepackage{tikz}
\usetikzlibrary{arrows.meta, positioning, fit, backgrounds, calc, shapes.multipart}
\usepackage{xcolor}

\def\BibTeX{{\rm B\kern-.05em{\sc i\kern-.025em b}\kern-.08em
    T\kern-.1667em\lower.7ex\hbox{E}\kern-.125emX}}
\begin{document}

\begin{titlepage}
\centering

\vspace*{2cm}

{\LARGE \textbf{Towards Realistic 3D Sonar Simulation}\par}

\vspace{1.5cm}

{\large
Youssef Attia, Davide Costa, Francesco Wanderlingh,\\
Filippo Campagnaro, Enrico Simetti
\par}

\vspace{1.5cm}

{\large
Accepted manuscript version\\
Accepted for publication in \textit{OCEANS 2026 Sanya}
\par}

\vspace{2cm}

\begin{minipage}{0.9\textwidth}
\small
\textbf{Copyright notice.}
\textcopyright{} 2026 IEEE. Personal use of this material is permitted.
Permission from IEEE must be obtained for all other uses, in any current
or future media, including reprinting/republishing this material for
advertising or promotional purposes, creating new collective works, for
resale or redistribution to servers or lists, or reuse of any copyrighted
component of this work in other works.
\end{minipage}

\vspace{1.5cm}

\begin{minipage}{0.9\textwidth}
\small
\textbf{Version notice.}
This is the accepted manuscript version of a paper accepted for
publication in \textit{OCEANS 2026 Sanya}. The final version of record
will be available through IEEE Xplore.
\end{minipage}

\vfill

{\small
arXiv version submitted under the arXiv.org perpetual, non-exclusive license.
\par}

\end{titlepage}

\setcounter{page}{1}
\title{Towards Realistic 3D Sonar Simulation\\
}

\author{
\IEEEauthorblockN{
Youssef Attia\textsuperscript{$\dagger$},
Davide Costa\textsuperscript{$\star$},
Francesco Wanderlingh\textsuperscript{$\dagger$$\ddagger$},
Filippo Campagnaro\textsuperscript{$\star$$\ddagger$},
Enrico Simetti\textsuperscript{$\dagger$$\ddagger$}
}
\IEEEauthorblockA{
\textsuperscript{$\dagger$}University of Genova, Italy
\textsuperscript{$\star$}University of Padova, Italy,\\
\textsuperscript{$\ddagger$}Interuniversity Research Center on Integrated Systems for the Marine Environment (ISME)\\
youssef.attia@edu.unige.it,
costadavid@dei.unipd.it,
francesco.wanderlingh@unige.it\\
filippo.campagnaro@unipd.it,
enrico.simetti@unige.it,
}
}
\maketitle

\begin{abstract}
As underwater robotics research increasingly addresses complex 3D perception and autonomous navigation, the fidelity of sonar simulation has become a key factor in algorithm development. Current simulation frameworks typically rely on geometry-driven rendering—approximating 3D sonar as an underwater equivalent to LiDAR—which fails to account for fundamental acoustic phenomena such as refraction, multi-path interference, and phase-dependent signal formation. This paper proposes a modular architecture for realistic 3D sonar simulation that integrates GPU-accelerated graphics engines with physically grounded acoustic propagation principles. We implement a volumetric 3D sonar model within the NVIDIA Isaac Sim environment, modeled after the Water Linked 3D-15 sensor, and integrate it into a comprehensive underwater simulation framework. The system is validated through a hardware-in-the-loop (HIL) configuration, where a modified FastLIO2 SLAM pipeline, executed on an NVIDIA Jetson Orin Nano, performs sensor fusion using synthetic 3D sonar, DVL, IMU, and pressure data. Finally, a qualitative comparison between simulated outputs and real-world data from harbor sheet-pile inspections is provided, characterizing the remaining sim-to-real gap and establishing a roadmap toward fully acoustics-driven volumetric sensing.
\end{abstract}


\section{Introduction and State of the Art}
\label{sec:SOA}
As underwater robotics has advanced, simulation has become an increasingly important tool for developing control, perception, and autonomy algorithms before field deployment. Yet the current simulator landscape remains heterogeneous: simulation frameworks differ in the level of physical fidelity they provide, in the realism of the environments they represent, in the sensor models they support, and in how effectively they enable sim-to-real transfer \cite{aldhaheri2025review}. This heterogeneity becomes especially visible in sonar simulation. Here, three partially overlapping lines of work have evolved in parallel: underwater robotics simulators designed to run in closed loop with control and perception stacks; sensor-oriented sonar rendering methods aimed at reproducing the appearance and statistical properties of real measurements; and underwater acoustic propagation models that capture wave phenomena more faithfully but are rarely embedded into mainstream robotics simulators. The gap addressed in this paper lies precisely at the intersection of these three directions.

Among general-purpose underwater simulation platforms relevant to sonar modeling, Gazebo, HoloOcean, Stonefish, and Isaac Sim constitute some of the most commonly used tools in underwater robotics research, although they differ substantially in fidelity, physical modeling, and intended application domain.
Gazebo, particularly in its classic and Ignition/Gazebo Sim variants, remains attractive because of its modularity, mature ROS integration, and suitability for control development and hardware-in-the-loop workflows. In underwater use, its capabilities are commonly extended through buoyancy and hydrodynamics plugins such as \emph{FreeBuoyancy}\footnote{\url{https://github.com/freefloating-gazebo/freefloating_gazebo}} and through broader frameworks such as DAVE, which provide marine environments, vehicle models, and sonar extensions \cite{dave2022}. Recent multibeam-sonar implementations in the Gazebo ecosystem go beyond simple depth rendering by incorporating beam geometry, point-scattering, reverberation surrogates, ambiguity effects, and GPU acceleration \cite{choi2021physicsbased,choi2025}. Nevertheless, the resulting pipeline is still generally centered on beam or ray interactions with geometric scene elements, followed by acoustically inspired post-processing, rather than on a full solution of underwater acoustic propagation.
\begin{figure}
    \centering
    \includegraphics[width=1.0\linewidth]{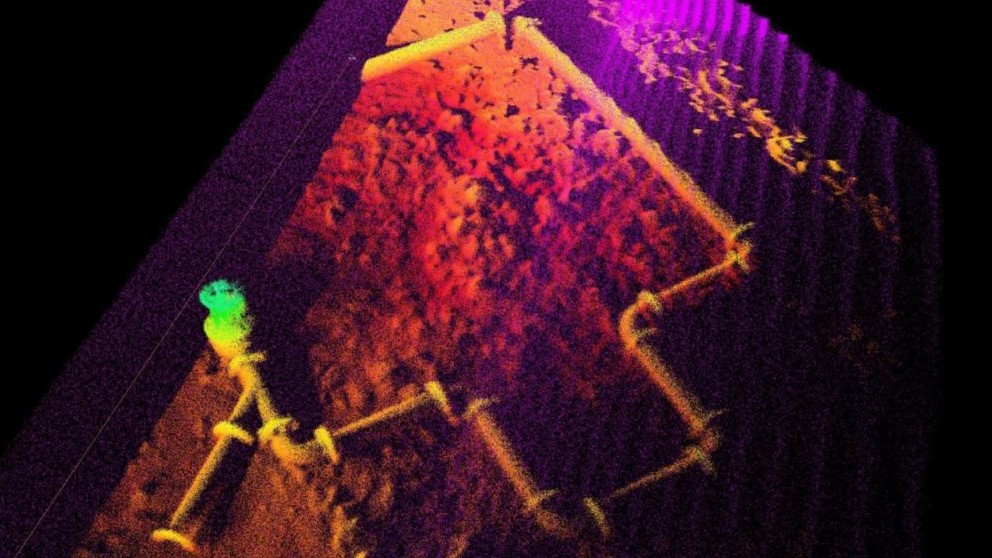}
    \caption{Output of the implemented 3D sonar simulation with hardware-in-the-loop SLAM, showing a point-cloud of an underwater pipeline structure.}
    \label{fig:Slamed}
\end{figure}
HoloOcean occupies a complementary point in the design space, prioritizing ease of use, rapid experimentation, and efficient rendering \cite{potokar_2022}. Built on an octree representation of the environment, it supports imaging, profiling, sidescan, and echosounder sonars, while introducing cluster-based multipath approximation, editable material dependence, and probabilistic noise models to improve realism \cite{Potokar22iros}. HoloOcean therefore offers an effective platform for algorithm development and synthetic-data generation, but its sonar pipeline remains an efficient approximate sensor model, not an explicit underwater acoustics solver.
Stonefish places stronger emphasis on physically grounded marine dynamics and on the tight coupling between underwater vehicle motion and sensing \cite{stonefish2019}. Its sensor suite includes forward-looking, multibeam, profiling, mechanical-scanning, and sidescan sonars, integrated with detailed hydrodynamics and marine robotics functionality. This makes Stonefish one of the most rigorous open-source options when perception must be studied together with vehicle and manipulator dynamics. However, even in this case, sonar outputs are still produced primarily through a rendering-oriented sensor model rather than through a general-purpose volumetric propagation framework.
At the GPU-accelerated end of the spectrum, Isaac Sim with the \emph{OceanSim} extension provides an especially compelling platform for fast underwater perception simulation and synthetic-data generation \cite{oceansim2025}. In its released imaging-sonar pipeline, scene geometry is queried through hardware-accelerated ray tracing, per-return intensity is modulated by reflectivity, incidence angle, and range attenuation, and the queried 3D returns are collapsed onto a 2D range--azimuth grid with additive and multiplicative speckle-like noise \cite{oceansim2025}. OceanSim is therefore a strong acoustically informed simulator for 2D imaging sonar and for large-scale GPU workflows, but it is not yet a volumetric 3D sonar model or a full underwater acoustic propagation solver.

Beyond platform-level simulators, a substantial body of work focuses specifically on sonar image formation. Cerqueira et al. proposed a hybrid rasterization/ray-tracing pipeline for real-time multi-device sonar simulation, computing transmission loss, reverberation, material-dependent response, and speckle-related effects while keeping the rendering cost tractable on GPU \cite{cerqueira2020rasterized}. Choi et al. introduced a physics-based point-scattering model for multibeam echosounders that reproduces realistic point-spread behavior and coherent speckle more faithfully than geometry-only rendering \cite{choi2021physicsbased}, and later extended this line with CUDA-accelerated ray-based sonar modeling inside a ROS--Gazebo framework, emphasizing directivity, interference, and bottom-grazing behavior \cite{choi2025}. Morency et al. similarly developed a high-fidelity forward-looking-sonar simulation environment aimed at sonar design evaluation and obstacle-detection studies \cite{morency2019fls}. Collectively, these works show that the field has already moved beyond naive depth-buffer rendering or purely cosmetic noise injection. Still, the dominant target remains realistic image synthesis for 2D sonar modalities, rather than volumetric 3D sensing with explicit propagation through the water column.

A second, partly separate line of work comes from underwater acoustics proper. Classical tools such as BELLHOP, KRAKEN, and TRACEO3D represent different families of propagation models, spanning geometric ray/beam tracing, normal-mode methods, and 3D ray-based formulations with coherent field reconstruction \cite{bellhop,porter1984numerical,traceo3d2017}. Comparative studies show that these models offer different trade-offs depending on frequency, bathymetry, boundary conditions, and computational budget \cite{wynnsimmonds2025propagation}. Their relevance is reinforced by the importance of environmental factors such as the sound-speed profile, whose spatial variation can materially alter refraction paths and therefore measurement geometry \cite{wu2024ssp}. These approaches are much better suited than conventional graphics renderers to represent refraction, boundary interactions, modal structure, and out-of-plane propagation, but they are typically too specialized or computationally demanding to function directly as drop-in real-time sensor models inside the closed-loop robotics simulators most commonly used by the community.

At the same time, downstream robotic perception is becoming increasingly three-dimensional. Earlier works exploited orthogonal imaging sonars or the fusion of forward-looking and profiling sonars to recover 3D structure and improve mapping quality \cite{mccconnell2020fusion,mccconnell2021predictive,joe2021sensorfusion}. More recent approaches use differentiable or volumetric sonar rendering for implicit reconstruction and novel-view synthesis, suggesting that inverse models of acoustic image formation are becoming central to learning-based underwater perception \cite{qadri2022implicit,sonarsplat2025}. Most importantly, recent results with compact 3D sonars demonstrate that volumetric acoustic sensing is moving from a specialized capability toward a practical sensing modality for mapping, simultaneous localization and mapping (SLAM), and motion planning \cite{burgul2025sonar,shrums}. This evolution raises the bar for simulation: a renderer tuned for plausible 2D imaging-sonar appearance may be sufficient for benchmarking image-level algorithms, yet it is often insufficient for studying how volumetric returns, multipath, and phase-related effects influence mapping and decision making.

These considerations highlight a persistent limitation in the current state of the art. Even with a collection of underwater simulators, and sophisticated acoustic propagation models in isolation, what is still missing is a practical bridge between them: a simulator architecture that preserves GPU acceleration and interoperability with modern robotics frameworks, while moving beyond purely geometry-driven sonar rendering toward volumetric 3D sensing, hybrid propagation, scattering-aware multipath, and phase-aware signal formation. In this sense, treating 3D sonar merely as an underwater analogue of RTX LiDAR is a useful engineering starting point, but not the final objective. The work that follows is motivated by this gap and outlines a path toward a more acoustically grounded 3D sonar simulation stack for underwater robotics.

The primary contributions of this work are as follows:
\begin{itemize}
    \item \textit{Architectural Framework}: We define a modular framework for volumetric 3D sonar simulation that combines geometry-based rendering with acoustic effects, including scattering and phase-aware signal formation, enabling more realistic modeling of sonar sensing.

    \item \textit{Simulator Implementation}: We implement this framework as an open-source 3D sonar simulator in Isaac Sim 5.1, extending the platform to generate volumetric point clouds consistent with real sensor characteristics and suitable for downstream robotics applications.

    \item \textit{SLAM Pipeline Integration for Validation}: To demonstrate that the simulated data is suitable for robotics workloads, we integrate the sonar model into an existing SLAM pipeline (FastLIO2) and extend it to incorporate DVL and pressure measurements. This integration is not meant to propose a new SLAM method, but to show that the simulator provides consistent, real-time data streams that support the development and testing of SLAM pipelines.

    \item \textit{Empirical Validation}: We evaluate the system through hardware-in-the-loop experiments and comparisons with real harbor-inspection data, assessing both the realism of the simulated measurements and their suitability for real-time processing on embedded platforms. A representative output of the resulting point-cloud generated via hardware-in-the-loop SLAM is shown in Fig.~\ref{fig:Slamed}.
\end{itemize}

The remainder of this paper is organized as follows. Section \ref{sec:ProposedSonarImpl} details the proposed architecture for 3D sonar simulation, focusing on volumetric sensing and hybrid propagation models. Section \ref{sec:3DSonarIsaac} describes the specific implementation of the 3D sonar as an RTX-based sensor within the Isaac Sim environment. Section \ref{sec:HILSim} presents the hardware in the loop validation and the integration of the sensor suite into a modified SLAM pipeline. Section \ref{sec:Sim2Real} discusses the comparison between simulated and real harbor inspection data, highlighting the current sim-to-real discrepancy. Finally, Section~\ref{sec:Conclusion} summarizes the main findings of this work.

\begin{figure*}[t]
\centering
\resizebox{0.75\textwidth}{!}{%
\begin{tikzpicture}[
    node distance=1cm and 1.2cm,
    core/.style={rectangle,
    draw=black,
    thick,
    fill=white,
    text width=2.5cm,
    align=center,
    minimum height=1.3cm},
    future/.style={rectangle,
    draw=gray,
    dashed,
    thick,
    fill=gray!5,
    text width=2.5cm,
    align=center,
    minimum height=1.3cm,
    text=gray},
    arrow/.style={-Stealth, thick},
    futureArrow/.style={-Stealth, thick, gray, dashed}
]

\node[core] (env) {Isaac Sim \\ Environment \\ \footnotesize(Scene Geometry) };

\node[core, right=of env] (sonar) {3D Sonar Model \\ \footnotesize (RTX LiDAR-style) };
\node[future, above=0.8cm of sonar] (prop) {Acoustic Solver \\ \footnotesize (BELLHOP/PE)};
\node[core, below=0.8cm of sonar] (signal) {Phase-Aware \\ Signal Formation};

\node[core, right=of sonar] (slam) {Modified \\ FastLIO2 Stack };
\node[core, below=0.8cm of slam] (fusion) {Sensor Fusion \\ \footnotesize (DVL, IMU, Pres.)};

\node[future, right=of slam] (nav) {3D Navigation \\ \& Mapping };
\node[future, right=of fusion] (gps) {GPS/LBL \\ Corrections };

\draw[arrow] (env) -- (sonar);
\draw[arrow] (sonar) -- (slam);
\draw[arrow] (signal) -- (sonar);
\draw[arrow] (fusion) -- (slam);

\draw[futureArrow] (prop) -- (sonar);
\draw[futureArrow] (slam) -- (nav);
\draw[futureArrow] (gps) -- (fusion);

\begin{scope}[on background layer]
    \node[draw=blue!20, fill=blue!2, dashed, inner sep=8pt, fit=(sonar) (signal), label={above:\footnotesize \textbf{Proposed Volumetric Model }}] {};
\end{scope}

\end{tikzpicture}
}
\caption{Block diagram of the proposed 3D sonar simulation and navigation architecture. The primary pipeline (solid black) consists of the Isaac Sim environment providing scene geometry, the volumetric RTX-based sonar model, and phase-aware signal formation. This information is integrated into a modified FastLIO2 SLAM stack that performs sensor fusion with IMU, DVL, and pressure data. Gray dashed blocks indicate future developments, such as integration with the BELLHOP propagation solver, Global Positioning System (GPS)/Long Baseline (LBL) corrections, and 3D navigation tasks.}
\label{fig:block_diagram}
\end{figure*}

\section{Proposed Sonar Implementation}
\label{sec:ProposedSonarImpl}
In this work, we do not present a complete implementation of a physics‑accurate sonar simulator. Instead, we outline a possible architecture that could extend current GPU‑accelerated sonar renderers towards a more acoustically driven, volumetric sensing model suitable for underwater robotics, and we discuss technological options that could be employed to realize such an architecture. The proposed system architecture and its integration with the navigation stack are illustrated in Fig.~\ref{fig:block_diagram}. The proposed design aims to preserve compatibility with existing 3D simulation frameworks while incorporating additional aspects of underwater acoustic propagation and scattering \cite{oceansim2025}.

\subsection{Volumetric 3D Sensing and Hybrid Propagation}
We envision a sonar sensor model that departs from the purely 2D imaging formulation (range–bearing) commonly employed in existing simulators and instead represents the sensor as a volumetric device. In this concept, the transmit–receive aperture is discretized over both azimuth and elevation, and the sensor output at each time step is a three‑dimensional distribution of backscattered intensity, for example organized as a voxel grid or as an azimuth–elevation–range point cloud. Such a volumetric formulation would be more naturally aligned with recent work on wide‑aperture imaging and 3D reconstruction from multibeam and forward‑looking sonar measurements \cite{choi2021physicsbased, choi2025}.

From a technological standpoint, the volumetric sensing model could be implemented on top of existing 3D digital‑twin pipelines (e.g., OpenUSD/Omniverse/Isaac‑based environments) by reusing their ray‑casting backends. In particular, hardware‑accelerated ray tracing via NVIDIA RTX and OptiX, or similar GPU ray‑tracing APIs, could be exploited to cast large numbers of rays in parallel through complex 3D scenes, as already demonstrated in other underwater acoustics and radar/SAR simulation contexts \cite{Willis_2020, Battle2023}.

On the propagation side, we propose a hybrid model in which geometric ray tracing is augmented with simplified water‑column physics. In this framework, rays would be traced through a spatially varying sound‑speed field and allowed to bend according to Snell’s law, thereby capturing first‑order refraction effects, basic shadowing and focusing phenomena in range‑ and depth‑dependent environments. The propagation kernel could be implemented in CUDA‑like frameworks (e.g., NVIDIA Warp or custom CUDA code) to update ray states on the GPU, following the approach already explored in parallel implementations of underwater acoustic ray tracers. The operating regime would be restricted to frequency and range conditions for which ray‑based methods provide an adequate approximation, thus avoiding the complexity of full parabolic‑equation or normal‑mode solvers, while still leaving the door open to future coupling with more advanced models (e.g., PE or FDTD) for specific scenarios \cite{Lazzarin2013, Jie_Weisong_Xiangjun_Xin_Malekian_2015, Oliveira2021}.

\subsection{Scattering, Multipath, Phase‑Aware Signal Formation}
At the interaction with the 3D scene, we expect a scattering model that goes beyond purely geometric first‑hit rendering while remaining computationally tractable. For each intersection between a ray and a surface element, the backscattered contribution could be computed as a function of the local incidence angle, the transmit–receive beam pattern, and material‑dependent reflection and scattering coefficients, potentially complemented by roughness statistics at the relevant wavelength to emulate speckle \cite{meng2021}. These scattering functions might be specified analytically, derived from established underwater scattering models, or learned from data using point‑based or differentiable ray‑tracing techniques that have recently been proposed for acoustic and radar applications \cite{CERQUEIRA2020101086}.

To approximate reverberation without incurring prohibitive cost, a limited number of secondary reflections and volumetric backscatter paths (for instance, from the water column or suspended targets) may be included, inspired by recursive ray‑acoustics and hybrid rasterized ray‑tracing approaches for sonar and SAR. The corresponding multi‑bounce ray paths could again be handled by RTX/OptiX‑style kernels on the GPU, with the maximum path depth treated as a tunable parameter that trades physical fidelity for runtime performance \cite{Wang2025}.

A further element of the proposed architecture is the use of phase‑aware signal formation. Instead of operating exclusively on intensities, the simulator could maintain complex‑valued pressure or baseband signals along each ray, accumulating phase according to propagation delay and frequency‑dependent attenuation. At the receiver, beams and volumetric cells would then be formed by coherent summation within each ray bundle, allowing the emergence of interference patterns, beam sidelobes and speckle statistics that are difficult to reproduce with purely intensity‑based image formation \cite{Bancel2021}. Providing access to time‑series data in addition to intensity volumes would facilitate the use of established signal‑processing techniques such as matched filtering, synthetic aperture processing and probabilistic 3D reconstruction, as demonstrated in recent sonar and SAR simulation frameworks that exploit GPU‑accelerated phase‑accurate rendering.

In a more advanced configuration, the simulation framework could also account for potential cross‑interference and coupling among co‑located acoustic sensors such as DVL, LBL, Ultra-Short Baseline (USBL), communication modems, echo sounders, and imaging sonars operating within the same frequency bands or overlapping time slots. This aspect, often neglected in idealized models, becomes relevant for realistic multi‑sensor deployments where mutual interference can manifest as transient echoes, spectrum leakage, or phase perturbations. A plausible implementation could represent each emitter as a temporally and spectrally characterized source, allowing the propagation engine to superimpose coherent or incoherent contributions across sensors and to expose the resulting distorted measurements to higher‑level control and navigation stacks. In this perspective, the proposed sonar rendering layer could be coupled with established underwater networking frameworks such as DESERT Underwater \cite{desert} and the corresponding ROS middleware \cite{rmwdesert}, where the first provides a realistic, protocol‑aware acoustic channel and network stack for underwater modems  and the second exposes that stack as a ROS2‑compliant transport for AUV and ROV acoustic control. Such an integration would enable end‑to‑end digital‑twin use cases in which vehicle guidance, cooperative scenarios and multi‑AUV mission logic are exercised against synthetic yet physically plausible sensor returns and communication impairments, supporting the design and testing of scheduling and sensor‑fusion strategies impacting on the sonar behavior.

Finally, this sonar model could be integrated into existing 3D digital‑twin pipelines by leveraging the same synthetic‑data infrastructure already used for underwater camera and imaging‑sonar rendering, for example Omniverse Replicator combined with GPU‑based parallel computing libraries such as NVIDIA Warp. In such a setup, the rendering engine would provide geometry, material metadata and acceleration structures, while the acoustic layer would implement the hybrid propagation, scattering and signal‑formation stages described above, thereby enabling a gradual evolution from “acoustically inspired rendering” towards a more explicitly acoustics‑driven sonar simulator.

\section{3D sonar as RTX LiDAR in Isaac Sim}
\label{sec:3DSonarIsaac}
The current public OceanSim release for Isaac Sim provides an imaging-sonar model, but not a 3D volumetric sonar sensor \cite{oceansim2025}. Moreover, its sonar formation pipeline is based on GPU-accelerated geometry queries with acoustic-inspired intensity and noise modeling, rather than an explicit propagation solver \cite{oceansim2025}. For this reason, we implement our 3D sonar in Isaac Sim as a continuation of OceanSim's infrastructure while leaving deeper integration with propagation tools such as BELLHOP as future work.
\begin{figure*}[!t]
    \centering
    \includegraphics[
        width=0.9\textwidth,
        trim=0.2cm 0.7cm 0.2cm 0.7cm,
        clip
    ]{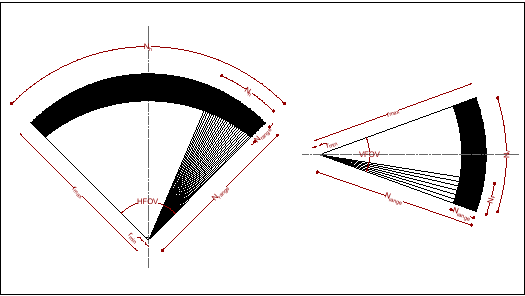}
    \caption{Geometric model of the 3D sonar, Left: Azimuth View;  Right: Elevation View}
    \label{fig:sonar_model}
\end{figure*}

In the 3D sonar model shown in Fig.~\ref{fig:sonar_model}, $\mathrm{HFOV}$ and $\mathrm{VFOV}$ denote the horizontal and vertical fields of view, while $r_{\max}$ and $r_{\min}$ denote the maximum and minimum sensing ranges of the sonar. Based on these quantities, the sonar sampling grid can be described in terms of the number of range bins, $N_{\mathrm{range}}$, the number of horizontal angular samples, $N_H$, and the number of vertical angular samples, $N_V$. These quantities are defined as
\begin{equation}
N_{\mathrm{range}}
\;=\;
\left\lfloor \dfrac{r_{\max}-r_{\min}}{\Delta r} \right\rfloor ,
\label{eq:Nrange}
\end{equation}
\begin{equation}
N_H
\;=\;
\left\lfloor \frac{\mathrm{HFOV}}{\Delta\theta_h} \right\rfloor ,
\label{eq:NH}
\end{equation}
\begin{equation}
N_V
\;=\;
\left\lfloor \frac{\mathrm{VFOV}}{\Delta\theta_v} \right\rfloor ,
\label{eq:NV}
\end{equation}
where $\Delta r$ is the range resolution, $\Delta\theta_h$ is the horizontal angular resolution, and $\Delta\theta_v$ is the vertical angular resolution.
In a solid-state LiDAR, $N_L$ denotes the number of elevation rows per scan; therefore, it is defined as $N_L = N_V + 1$. The number of rays per elevation line is $N_R = N_H + 1$. Finally, the number of emitters (beams) per scan, $N_E$, is the total number of grid points in the firing pattern, i.e., $N_E = N_L \, N_R$.
The values were extracted from the Water Linked 3D Sonar datasheet\footnote{\label{fn:waterlinked_sonar}\url{https://www.waterlinked.com/datasheets/sonar-3D-15}}, and the Isaac Sim sensor definition and configuration files are open-sourced\footnote{\label{fn:opensource_repo}\url{https://github.com/GRAAL-Lab/IsaacSim_Underwater}}.

\section{Hardware in the Loop Simulation}
\label{sec:HILSim}
The purpose of this section is to assess whether the proposed SLAM pipeline can operate in real time on embedded hardware when driven by the sensor data generated in simulation. To this end, we adopted a hardware in the loop (HIL) setup in which the underwater scenario and sensor suite are simulated in Isaac Sim 5.1, while the SLAM stack is executed on an external NVIDIA Jetson Orin Nano. In this configuration, the Jetson receives the simulated measurements and processes them as it would in a real deployment.

The simulated scenario consists of a BlueROV2 operating over a nearly flat seabed with a pipe structure representative of real underwater environments. The simulated sensor suite includes the proposed 3D sonar, a Doppler Velocity Log (DVL), a 6-axis Inertial Measurement Unit (IMU) using only accelerometer and gyroscope measurements, and a pressure sensor.

FastLIO2 \cite{fastlio2} was selected after comparison with alternative approaches, including the KISS-ICP SLAM framework used in \cite{burgul2025sonar}. Unlike FastLIO2, KISS-ICP relies only on sonar data and does not perform sensor fusion, which makes it less suitable for the multi-sensor underwater configuration considered in this work. A fork of the open-source FastLIO2 repository was therefore identified and modified to exploit the computational capabilities of the Jetson Orin Nano, which can be readily integrated into the BlueROV2 \footnote{\url{https://github.com/OmerMersin/FAST_LIO_GP}}. Additional modifications were introduced to improve runtime performance, support the integration of the underwater sensor suite, and extend the Iterated Kalman Filter to incorporate DVL and pressure measurements together with sonar and IMU data.

Figure~\ref{fig:Slamed} shows the result of this validation, in which the BlueROV2 successfully mapped an area of approximately \qty{25}{\metre\squared}. These results indicate that the proposed simulation and perception pipeline can already support real-time operation on embedded hardware for representative underwater mapping tasks.

\section{From Simulation to Real World}
\label{sec:Sim2Real}
\subsection{Harbor sheet-pile inspection as a motivating use case}
\begin{figure}[!htbp]
    \centering
    \includegraphics[width=\linewidth]{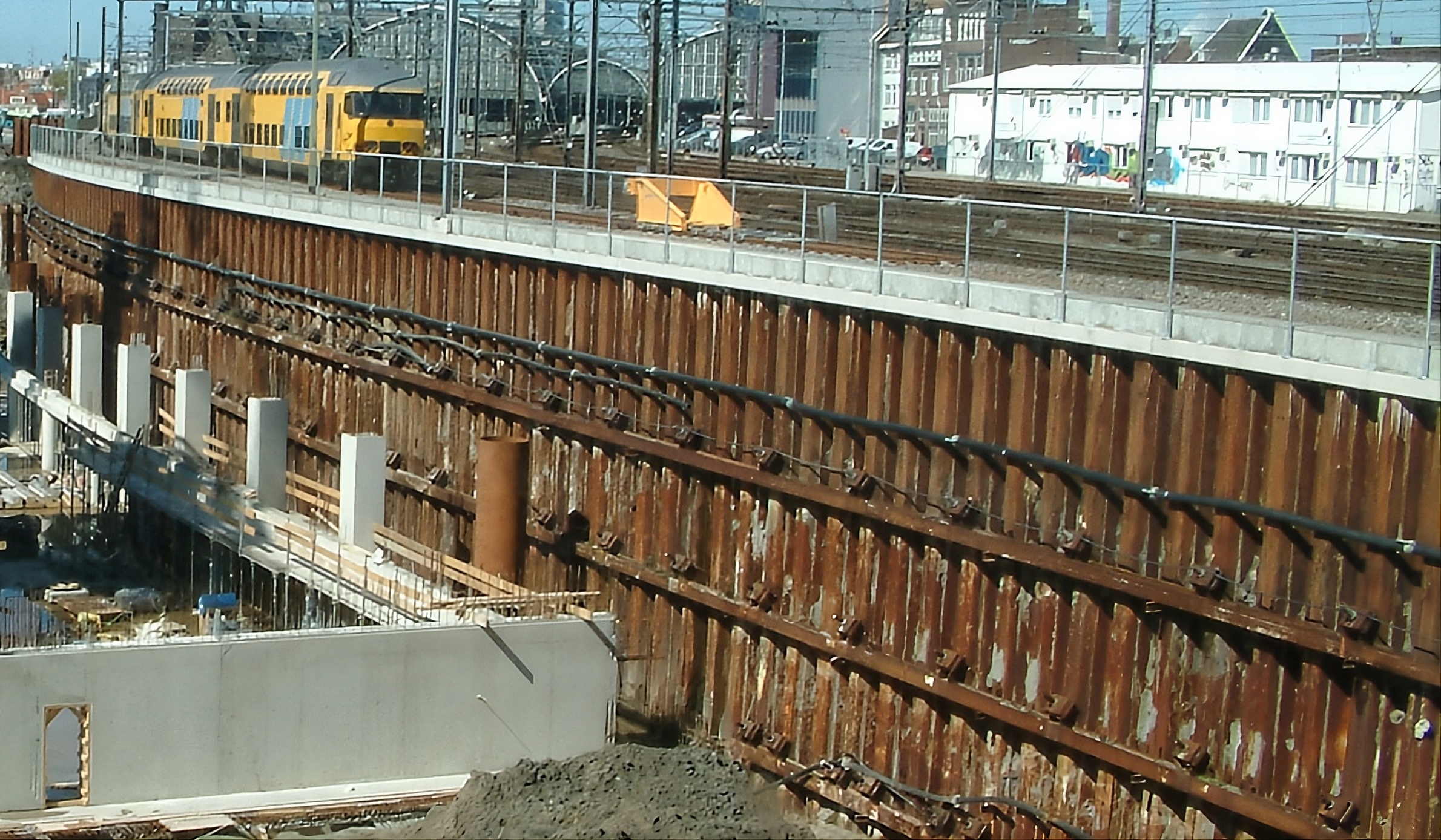}
    \caption{Example of a steel sheet-pile wall, illustrating the interlocking vertical elements typical of quay-wall structures in construction sites and harbors \cite{fons2005sheetpile}.}
    \label{fig:sheetpile_example}
\end{figure}

Steel sheet-pile walls, such as the one shown in Fig.~\ref{fig:sheetpile_example}, are widely used  as quay walls and retaining structures in ports, where a row of interlocking vertical steel elements forms a nearly continuous waterfront barrier. Their underwater portions require periodic inspection because degradation mechanisms such as corrosion, coating loss, local deformation, impact damage, and scour can reduce structural integrity over time \cite{mucke2015hamburg}.

In harbor waters, optical inspection often becomes unreliable because turbidity, suspended particles, and limited illumination severely degrade image quality; in extreme cases, divers must rely more on tactile assessment than on vision \cite{belcher2000turbid,torrey2019underwatercorrosion}. For this reason, 3D sonar is an attractive inspection modality. Recent quay wall surveys have shown that acoustic systems can generate detailed 3D models of vertical waterfront structures, detect bent sheet piles, distorted steel plates, cracks, erosion, and other weak points, and support maintenance planning even in confined harbor environments \cite{mucke2015hamburg,silden2024quaywall}. This makes harbor sheet-pile inspection a representative benchmark for evaluating realistic underwater perception pipelines.

\subsection{Why the next step must be acoustics-based}
\begin{figure*}[!htbp]
    \centering
    \includegraphics[width=\textwidth]{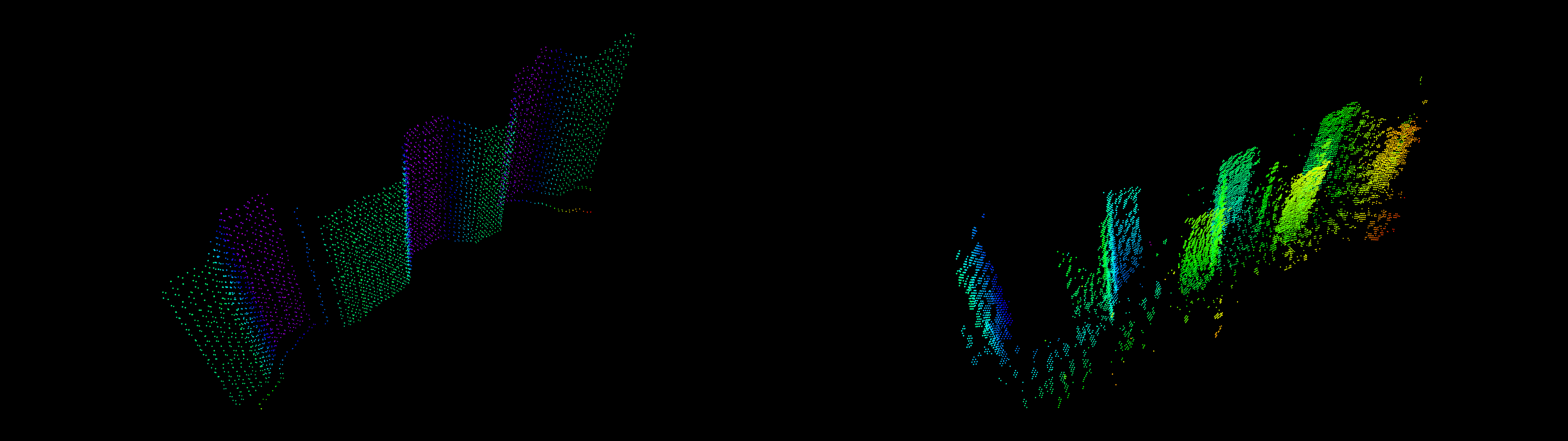}
    \caption{Harbor sheet-pile point cloud: comparison between the point cloud generated by the proposed Isaac-Sim-based 3D sonar model (left) and a real 3D sonar acquisition collected at a harbor test site (right). The real measurement is markedly sparser and more cluttered, with non-uniform density and partial occlusions that are not fully captured by a geometry-only simulation.}
    \label{fig:sim_real_comparison}
\end{figure*}
To assess the current limitations of the proposed simulator, a harbor sheet-pile scenario was reproduced in Isaac Sim using the proposed 3D sonar model, and the resulting synthetic point cloud was compared with a real 3D sonar acquisition collected at a harbor test site. This comparison is shown in Fig.~\ref{fig:sim_real_comparison}.

The comparison highlights a practical sim-to-real gap. The current RTX LiDAR based implementation is effective for sensor emulation, SLAM integration, and HIL validation because it preserves the sensor field of view and the dominant scene geometry. However, the real harbor acquisition is visibly different from the synthetic one. The returns are sparser and less uniform, some surfaces appear partially missing, and the scene contains clutter that is not reproduced by a simple first-intersection model.

On the other hand, the state of the art simulators are not completely non-physical. HoloOcean and OceanSim already incorporate acoustically informed elements such as material dependence, incidence-angle modulation, attenuation, multipath approximation, and stochastic speckle and noise modeling \cite{Potokar22iros,oceansim2025}. Nevertheless, their released sonar pipelines remain primarily geometry-driven and focused on 2D imaging sonar formation rather than 3D volumetric sensing. For inspection and SLAM in cluttered harbor environments, the residual discrepancy suggests that additional propagation-aware modeling is needed, including sound-speed-driven refraction, broader multipath and reverberation effects, and more explicit scattering and sensor formation physics. Such a model would complement, rather than replace, the current Isaac Sim implementation by narrowing the sim-to-real gap for inspection, SLAM, and change-detection tasks.

\section{Conclusion}
\label{sec:Conclusion}
This work presents a step toward realistic 3D sonar simulation for underwater robotics. The main contribution is a complete GPU-accelerated underwater simulation framework in Isaac Sim 5.1 that integrates vehicle dynamics, ROS 2 interfacing, and a heterogeneous sensor suite including a 3D sonar, a 2D imaging sonar, an underwater camera, a DVL, a pressure sensor, and an IMU. Within this framework, the proposed 3D sonar is currently implemented as an RTX LiDAR approximation inspired by the Water Linked 3D-15, enabling volumetric point cloud generation and integration into SLAM pipelines. In parallel, the paper outlines a broader architecture for future sonar simulation that moves beyond geometry-based rendering toward a more acoustically grounded formulation based on volumetric sensing, hybrid propagation, scattering-aware multipath, and phase-aware signal formation.

The results obtained with the HIL setup confirm that the framework is suitable for closed-loop testing with embedded computing. Running the simulator in Isaac Sim while executing SLAM on the Jetson Orin Nano shows that the proposed approach is practical for evaluating underwater perception and navigation pipelines under realistic computational constraints.

At the same time, the comparison between simulated and real acquisitions highlights a remaining sim-to-real gap. In particular, the real data exhibit sparsity, clutter, partial surface visibility, and other effects that are not fully reproduced by the current first-intersection geometry-based model. This observation confirms that the next step should focus on stronger acoustics-based modeling.

\section*{Acknowledgment}
The results of this research have been obtained in part within the framework of the National PhD Program in Robotics and Intelligent Machines. Mr. Attia's scholarship is co-funded by Graal Tech Srl and under Ministerial Decree 630 of 24/04/2024, within the framework of the National Recovery and Resilience Plan (PNRR), Mission 4, Component 2, 'From Research to Business' – Investment 3.3, 'Introduction of innovative PhD programs that address the innovation needs of companies and promote the hiring of researchers by enterprises.' and in part by the European Union - NextGenerationEU and by the Ministry of University and Research (MUR), National Recovery and Resilience Plan (NRRP), Mission 4, Component 2, Investment 1.5, project “RAISE - Robotics and AI for Socio-economic Empowerment” (ECS00000035). Enrico Simetti is part of RAISE personnel.
\bibliography{oceans_sonar}
\bibliographystyle{IEEEtran}

\vspace{12pt}
\end{document}